\documentclass[runningheads]{llncs}
\usepackage{url}
\usepackage{graphicx} 
\usepackage{siunitx}
\usepackage{verbatim}
\usepackage[misc]{ifsym}
\usepackage{graphicx}
\usepackage{amsmath}
\usepackage{pifont}
\usepackage{algorithm2e}
\usepackage{caption}
\usepackage{subcaption}
\usepackage[labelformat=simple]{subcaption}

\usepackage{multirow,multicol,ctable}
\usepackage{amsmath,amssymb,amsfonts}
\usepackage{algorithmic}
\usepackage{textcomp} 
\usepackage{booktabs}
\usepackage{subcaption}
\usepackage{amsfonts}
\usepackage{multirow,multicol,ctable,arydshln}
\usepackage{xcolor}
\usepackage{float}
\usepackage{makecell}
\usepackage{color}
\begin{document}
\setlength{\textfloatsep}{5pt}
\title{A Novel Multi-Task Model Imitating Dermatologists for Accurate Differential Diagnosis of Skin Diseases in Clinical Images}
\titlerunning{DermImitFormer for Skin Disease Differential Diagnosis in Clinical Images}
%
\author{Yan-Jie Zhou\inst{1,2}\textsuperscript{(\Letter)} \and Wei Liu \inst{1,2} \and Yuan Gao \inst{1,2} \and Jing Xu\inst{1,2} \and Le Lu\inst{1} \and \\ Yuping Duan\inst{3} \and Hao Cheng \inst{4} \and Na Jin \inst{4} \and Xiaoyong Man \inst{5} \and \\ Shuang Zhao \inst{6} \and Yu Wang\inst{1}\textsuperscript{(\Letter)} }
\authorrunning{Y.J. Zhou \textit{et al.}}

%
\institute{DAMO Academy, Alibaba Group \and
Hupan Lab, Hangzhou, China \\
\email{zhouyanjie.zyj@alibaba-inc.com, Flimanadam@gmail.com} \and
School of Mathematical Sciences, Beijing Normal University, Beijing, China \and
Sir Run Run Shaw Hospital, Hangzhou, China \and
The Second Affiliated Hospital Zhejiang University School of Medicine, China \and
Xiangya Hospital Central South University, Changsha, China
}

%
%
\maketitle      
%
\begin{abstract}
Skin diseases are among the most prevalent health issues, and accurate computer-aided diagnosis methods are of importance for both dermatologists and patients. However, most of the existing methods overlook the essential domain knowledge required for skin disease diagnosis. A novel multi-task model, namely \textbf{DermImitFormer}, is proposed to fill this gap by imitating dermatologists’ diagnostic procedures and strategies. Through multi-task learning, the model simultaneously predicts body parts and lesion attributes in addition to the disease itself, enhancing diagnosis accuracy and improving diagnosis interpretability. The designed lesion selection module mimics dermatologists’ zoom-in action, effectively highlighting the local lesion features from noisy backgrounds. Additionally, the presented cross-interaction module explicitly models the complicated diagnostic reasoning between body parts, lesion attributes, and diseases. To provide a more robust evaluation of the proposed method, a large-scale clinical image dataset of skin diseases with significantly more cases than existing datasets has been established. Extensive experiments on three different datasets consistently demonstrate the state-of-the-art recognition performance of the proposed approach.
 
\end{abstract}
\section{Introduction}

As the largest organ in the human body, the skin is an important barrier protecting the internal organs and tissues from harmful external substances, such as sun exposure, pollution, and microorganisms \cite{kshirsagar2022deep,li2020deep}. In recent years, the increasing number of deaths by skin diseases has aroused widespread public concern \cite{rogers2015incidence,siegel2022cancer}. Due to the complexity of skin diseases and the shortage of dermatological expertise resources, developing an automatic and accurate skin disease diagnosis framework is of great necessity.

Among non-invasive skin imaging techniques, dermoscopy is currently widely used in the diagnosis of many skin diseases \cite{binder1995epiluminescence,kittler2002diagnostic}, but it is technically demanding and not necessary for many common skin diseases. Clinical images, on the contrary, can be easily acquired through consumer-grade cameras, increasingly utilized in teledermatology, but their diagnostic value is underestimated. Recently, deep learning-based methods have received great attention in clinical skin disease image recognition and achieved promising results \cite{IAS,DSN,liu2020deep,sun2016benchmark,wu2020learning,yang2019self,zhang2019medical,zhang2019attention}. Sun \textit{et al.} \cite{sun2016benchmark} released a clinical image dataset of skin diseases, namely SD-198, containing 6,584 images from 198 different categories. The results demonstrate that deep features from convolutional neural networks (CNNs) outperform hand-crafted features in exploiting structural and semantic information. Gupta \textit{et al.} \cite{DSN} proposed a dual stream network that employs class activation maps to localize discriminative regions of the skin disease and exploit local features from detected regions to improve classification performance.

\begin{figure}[!t]
    \centering
    \includegraphics[width=1.0\linewidth]{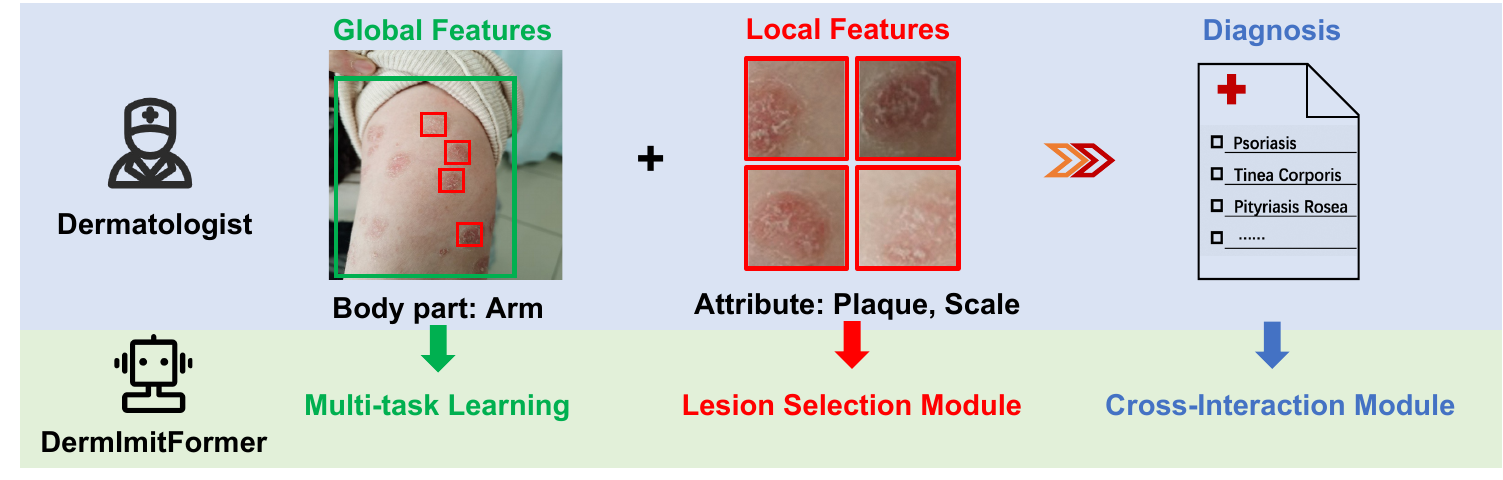}
    \caption{The relationship between dermatologists’ diagnostic procedures and our proposed model (best viewed in color).}
    \label{fig:figure1}
\end{figure}

Although these approaches have achieved impressive results, most of them neglect the domain knowledge of dermatology and lack interpretability in diagnosis basis and results. In a typical inspection, dermatologists give an initial evaluation with the consideration of both global information, e.g. body part, and local information, e.g. the attributes of skin lesions, and further information including the patient's medical history or additional examination is required to draw a diagnostic conclusion from several possible skin diseases. Recognizing skin diseases from clinical images presents various challenges that can be summarized as follows: (1) Clinical images taken by portable electronic devices (e.g. mobile phones) often have cluttered backgrounds, posing difficulty in accurately locating lesions. (2) Skin diseases exhibit high intra-class variability in lesion appearance, but low inter-class variability, thereby making discrimination challenging. (3) The diagnostic reasoning of dermatologists is empirical and complicated, which makes it hard to simulate and model.

To tackle the above issues and leverage the domain knowledge of dermatology, we propose a novel multi-task model, namely \textbf{DermImitFormer}. The model is designed to imitate the diagnostic process of dermatologists (as shown in Fig. \ref{fig:figure1}), by employing three distinct modules or strategies. Firstly, the multi-task learning strategy provides extra body parts and lesion attributes predictions, which enhances the differential diagnosis accuracy with the additional correlation from multiple predictions and improves the interpretability of diagnosis with more supporting information. Secondly, a lesion selection module is designed to imitate dermatologists' zoom-in action, effectively highlighting the local lesion features from noisy backgrounds. Thirdly, a cross-interaction module explicitly models the complicated diagnostic reasoning between body parts, lesion attributes, and diseases, increasing the feature alignments and decreasing gradient conflicts from different tasks. Last but not least, we build a new dataset containing 57,246 clinical images. The dataset includes 49 most common skin diseases, covering 80\% of the consultation scenarios, 15 body parts, and 27 lesion attributes, following the International League of Dermatological Societies (ILDS) guideline \cite{nast20162016}. 

The main contributions can be summarized as follows: (1) A novel multi-task model DermImitFormer is proposed to imitate dermatologists’ diagnostic processes, providing outputs of diseases, body parts, and lesion attributes for improved clinical interpretability and accuracy. (2) A lesion selection module is presented to encourage the model to learn more distinctive lesion features. A cross-interaction module is designed to effectively fuse three different feature representations. (3) A large-scale clinical image dataset of skin diseases is established, containing significantly more cases than existing datasets, and closer to the real data distribution of clinical routine. More importantly, our proposed approach achieves the leading recognition performance on three different datasets.

\section{Method}

\begin{figure}[!t]
    \centering
    \includegraphics[width=1.0\linewidth]{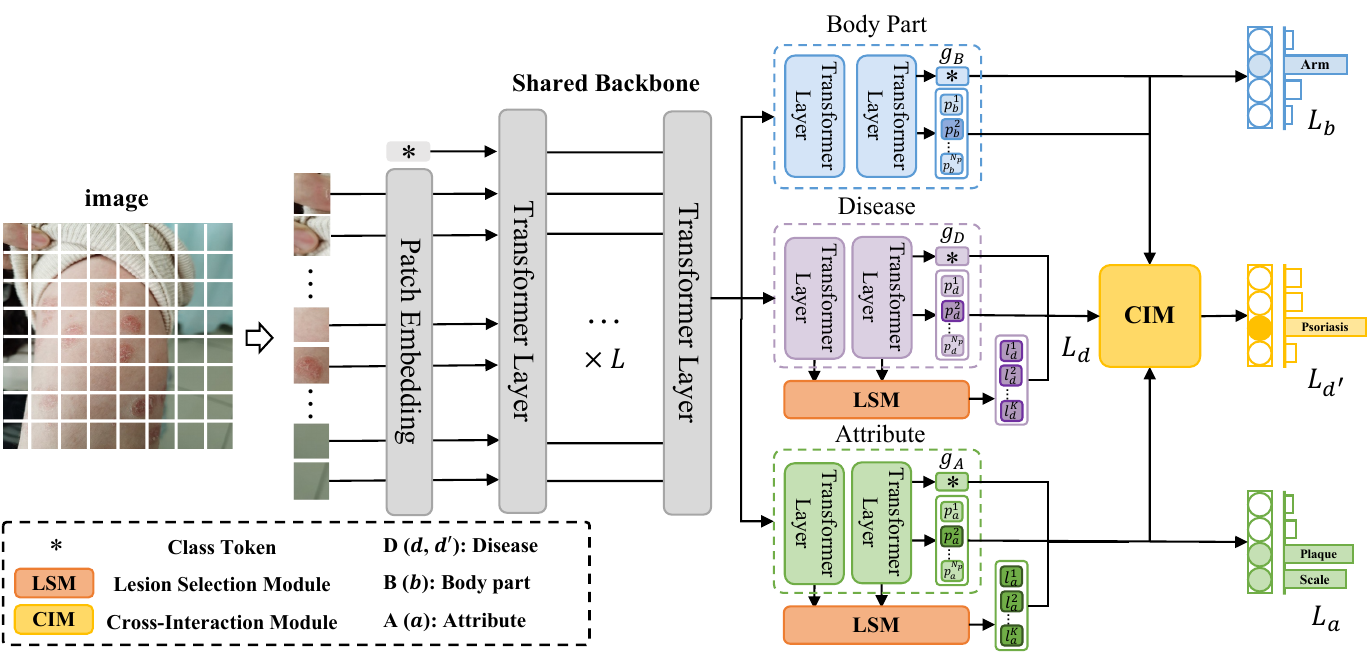}
    \caption{The overall architecture of the multi-task imitation model (DermImitFormer) with shared backbone and task-specific heads.}
    \label{fig:figure2}
\end{figure}

The architecture of the proposed multi-task model DermImitFormer is shown in Fig. \ref{fig:figure2}. It takes the clinical image as input and outputs the classification results of skin diseases, body parts, and attributes in an end-to-end manner. During diagnostic processes, dermatologists consider local and global contextual features of the entire clinical image, including shape, size, distribution, texture, location, etc. To effectively capture these visual features, we use the vision transformer (ViT) \cite{dosovitskiy2020image} as the shared backbone. Three separate task-specific heads are then utilized to predict diseases, body parts, and attributes, respectively, with each head containing two independent ViT layers. In particular, in the task-specific heads of diseases and attributes, the extracted features of each layer are separated into the image features and the patch features. These two groups of features are fed into the lesion selection module (LSM), to select the most informative lesion tokens. Finally, the feature representations of diseases, body parts, and attributes are delivered to the cross-interaction module (CIM) to generate a more comprehensive representation for the final differential diagnosis.
\vspace{-1em}
\subsubsection{Shared Backbone}

Following the ViT model, an input image $X$ is divided to $N_p$ squared patches $\{x_{n}, n\in \{1,2,..., N_p\}\}$, where $N_p = (H\times W) / P^{2}$, $P$ is the side length of a squared patch, $H$ and $W$ are the height and width of the image, respectively. Then, the patches are flattened and linearly projected into patch tokens with a learnable position embedding, denoted as $t_{n}, n\in \{1,2,..., N_p\}$. Together with an extra class token $t_{0}$, the network inputs are represented as ${t_{n}}\in \mathbb{R}^{D}, n\in \{0,1,...,N_p\}$ with a dimension of $D$. Finally, the tokens are fed to $L$ consecutive transformer layers to obtain the preliminary image features.
\vspace{-1em}
\subsubsection{Lesion Selection Module}

As introduced above, skin diseases have high variability in lesion appearance and distribution. Thus, it requires the model to concentrate on lesion patches so as to describe the attributes and associated diseases precisely. The multi-head self-attention (MHSA) block in ViT generates global attention,  weighing the informativeness of each token. Inspired by \cite{wang2021feature}, we introduce a lesion selection module (LSM), which guides the transformer encoder to select the tokens that are most relevant to lesions at different levels. 
Specifically, for each attention head in MHSA blocks, we compute the attention matrix $\boldsymbol{A}^{m}=\text{Softmax}(\mathcal{Q}\mathcal{K}^{T}/\sqrt{D})\in \mathbb{R}^{(N_p+1)\times (N_p+1)}$, where $m \in \left\{1,2,..., N_{h}\right\}$, $N_{h}$ denoting the number of heads, $\mathcal{Q}$ and $\mathcal{K}$ the $Query$ and $Key$ representations of the block inputs, respectively. The first row calculates the similarities between the class token and each patch token. As the class token is utilized for classification, the higher the value, the more informative each token is. We apply softmax to the first row and the first column of $\boldsymbol{A}^{m}$, denoted as $a_{0, n}^{m}$ and $a_{n, 0}^{m}, n\in \{1,2,..., N_p\}$, representing the attention scores between the class token and other tokens:
\begin{equation}
    a_{0,n}^m = \frac{{{e^{A_{0,n}^m}}}}{{\sum\limits_{i = 1}^{{N_p}} {{e^{A_{0,i}^m}}} }}, \quad  
    a_{n,0}^m = \frac{{{e^{A_{n,0}^m}}}}{{\sum\limits_{i = 1}^{{N_p}} {{e^{A_{i,0}^m}}} }}, \quad 
    {s_n} = \frac{1}{{{N_h}}}\sum\limits_{m = 1}^{{N_h}} {a_{0,n}^m \cdot a_{n,0}^m}
\end{equation}
The mutual attention score $s_{n}$ is calculated across all attention heads. Thereafter, we select the top $K$ tokens according to $s_{n}$ for two task heads as
$\boldsymbol{l}_{d}^{k}$ and $\boldsymbol{l}_{a}^{k}, k\in \{1,2,...,K\}$.
\vspace{-1em}
\subsubsection{Cross-Interaction Module}

\begin{figure}[t]
    \centering
    \includegraphics[width=1.0\linewidth]{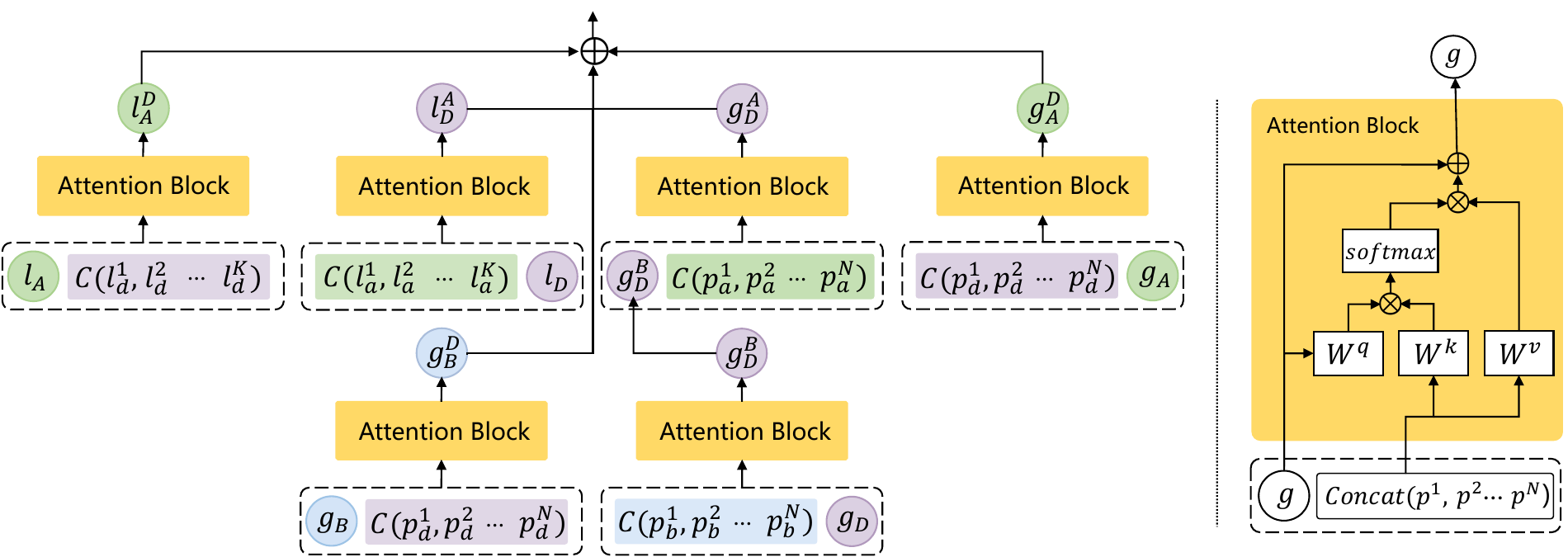}
    \caption{Schematic of cross-interaction module.}
    \label{fig:figure3}
\end{figure}

A diagnostic process of skin diseases takes multiple visual information into account, which is relatively complicated and difficult to model in an analytical way. Simple fusion operations such as concatenation are insufficient to simulate the diagnostic logic. Thus, partially inspired by \cite{xu2022remixformer}, the CIM is designed to learn complicated correlations between disease, body part, and attribute. 
The detailed module schematic is shown in Fig.~\ref{fig:figure3}. Firstly, the features of body-part and disease are integrated to enhance global representations by a cross-attention block. For example, the fusion between the class token of disease and patch tokens of body-part is:
\begin{equation}
z_b = LN(GAP({\boldsymbol{p}_b^1, \boldsymbol{p}_b^2, ......, \boldsymbol{p}_b^{N_{p}}}))
\end{equation}
\begin{equation}
\mathcal{Q} = LN\left(\boldsymbol{g}_{D}\right)\boldsymbol{W}_{BD}^\mathcal{Q}, \quad \mathcal{K} =  \boldsymbol{z}_b \boldsymbol{W}_{BD}^\mathcal{K}, \quad \mathcal{V} = \boldsymbol{z}_b\boldsymbol{W}_{BD}^\mathcal{V} \\
\end{equation}
\begin{equation}
\boldsymbol{g}^{B}_{D} = LN\left(\boldsymbol{g}_{D}\right) + \text{linear}(\text{softmax}(\frac{\mathcal{Q}\mathcal{K}^{T}}{\sqrt{F/{N_{h}}}})\mathcal{V})
\end{equation}
where $\boldsymbol{g}_B,\ \boldsymbol{g}_D$ are the class token, $\boldsymbol{p}_b^i,\ \boldsymbol{p}_d^i, i \in \left\{1,2,..., N_{p}\right\}$ the corresponding patch tokens. $GAP$ and $LN$ denote the global average pooling and layer normalization, respectively. $\boldsymbol{W}_{BD}^{\mathcal{Q}}, \boldsymbol{W}_{BD}^{\mathcal{K}},\boldsymbol{W}_{BD}^{\mathcal{V}}\in \mathcal{R}^{F\times F}$ denote learnable parameters. $F$ denotes the dimension of features. $\boldsymbol{g}^{D}_{B}$ is computed from the patch tokens of disease and the class token of body-part in the same fashion. Similarly, we can obtain the fused class tokens ($\boldsymbol{g}^{D}_{A}$ and  $\boldsymbol{g}^{A}_{D}$) and the fused local class tokens ($\boldsymbol{l}^{D}_{A}$ and  $\boldsymbol{l}^{A}_{D}$) between attribute and disease. Note that the disease class token $\boldsymbol{g}_D$ is replaced by $\boldsymbol{g}^{B}_{D}$ in the later computations, and local class tokens $\boldsymbol{l}_A$ and $\boldsymbol{l}_D$ in Fig.~\ref{fig:figure3} are generated by $GAP$ on selected local patch tokens from LSM. Finally, these mutually enhanced features from CIM are concatenated together to generate more accurate predictions of diseases, body parts, and attributes. 

\vspace{-1em}
\subsubsection{Learning and Optimization}
We argue that joint training can enhance the feature representation for each task. Thus, we define a multi-task loss as follows:
\vspace{-0.5em}
\begin{equation}
   \mathcal{L}_{x}  = - {\frac{1}{N_s} {\sum\limits_{i = 1}^{{N_s}} {\sum\limits_{j = 1}^{{n_x}} {{y_{ij} \log \left(p_{ij}\right)} }}}}, \quad
   x \in \{d, d'\}
\end{equation}
\vspace{-0.8em}
\begin{equation}
\mathcal{L}_{h}  = - {\frac{1}{N_s} {\sum\limits_{i = 1}^{{N_s}} {\sum\limits_{j = 1}^{{n_h}} {{y_{ij} \log \left(p_{ij}\right)} + {\left(1-y_{ij}\right) \log \left(1-p_{ij}\right)} }}}}, \quad h \in \{a, b\}
\end{equation}
\vspace{-0.8em}
\begin{equation}
\mathcal{L} = \mathcal{L}_{d} + \mathcal{L}_{d'} + \mathcal{L}_{b} +  \mathcal{L}_{a}
\label{equ: equ5}
\end{equation}
where ${N_s}$ denotes the number of samples, ${n_x}, {n_h}$ the number of classes for each task, and $p_{ij}, \ y_{ij}$ the prediction and label, respectively. Notably, body parts and attributes are defined as multi-label classification tasks, optimized with the binary cross-entropy loss, as shown in Eq. 6. The correspondence of $x$ and $h$ is shown in Fig. \ref{fig:figure2}.

\begin{table}[t]
\centering
\setlength{\tabcolsep}{1.5mm}
\caption{\label{tab:ablation1} Ablation study for DermImitFormer on Derm-49 dataset. D, B, and A denote the task-specific head of diseases, body parts, and attributes, respectively.}
    \begin{tabular}{cccccccc}
    \toprule
    \multirow{2}{*}{Dimension} & \multirow{2}{*}{LSM} & \multicolumn{2}{c}{Fusion} & \multicolumn{3}{c}{F1-score (\%)} & \multicolumn{1}{c}{Accuracy(\%)}\\
    \cmidrule(lr){3-4} \cmidrule(lr){5-7} \cmidrule(lr){8-8}
     & & Concat & CIM & Disease & Body part & Attribute & Disease \\
    \midrule
    D               &           & &           & 76.2 & -     & -     & 80.4 \\
    D               & \ding{51} & &           & 77.8 & -     & -     & 82.0 \\
    D + B           & \ding{51} & \ding{51} &           & 78.1 & 85.0 & -     & 82.4 \\
    D + A           & \ding{51} & \ding{51} &           & 78.4 & -     & 68.7 & 82.6 \\
    D + B + A       & \ding{51} & \ding{51} &           & 79.1 & 85.1 & 69.0 & 82.9 \\
    D + B + A       & \ding{51} & &\ding{51} & \textbf{79.5} & \textbf{85.9} & \textbf{70.4} & \textbf{83.3} \\
    \bottomrule
    \end{tabular}
\end{table}

\section{Experiment}

\subsubsection{Datasets}
The proposed DermImitFormer is evaluated on three different clinical skin image datasets including an in-house dataset and two public benchmarks. (1) \textbf{Derm-49:} We establish a large-scale clinical image dataset of skin diseases, collected from three cooperative hospitals and a teledermatology platform. The 57,246 images in the dataset were annotated with the diagnostic ground truth of skin disease, body parts, and lesions attributes from the patient records. We clean up the ground truth into 49 skin diseases, 15 body parts, and 27 lesion attributes following the ILDS guidelines \cite{nast20162016}. (2) \textbf{SD-198} \cite{sun2016benchmark}\textbf{:} It is one of the largest publicly available datasets in this field containing 198 skin diseases and 6,584 clinical images collected through digital cameras or mobile phones. (3) \textbf{PAD-UFES-20} \cite{pacheco2020impact}\textbf{:} The dataset contains 2,298 samples of 6 skin diseases. Each sample contains a clinical image and a set of metadata with labels such as diseases and body parts. 
\vspace{-1.5em}
\subsubsection{Implementation Details}
The DermImitFormer is initialized with the pre-trained ViT-B/16 backbone and optimized with SGD method (initial learning rate 0.003, momentum 0.95, and weight decay $10^{-5}$) for 100 epochs on 4 NVIDIA Tesla V100 GPUs with a batch size of 96. We define the input size i.e. $H=W=384$ that produces a total of 576 spatial tokens i.e. $N_{p}=576$ for a ViT-B backbone. $K$ = 24 in the LSM module. For data augmentation, we employed the Cutmix \cite{yun2019cutmix} with a probability of 0.5 and Beta(0.3, 0.3) during optimization. We adopt precision, recall, F1-score, and accuracy as the evaluation metrics.
\vspace{-1.5em}
\begin{figure}[t]
    \centering
    \includegraphics[width=1.0\linewidth]{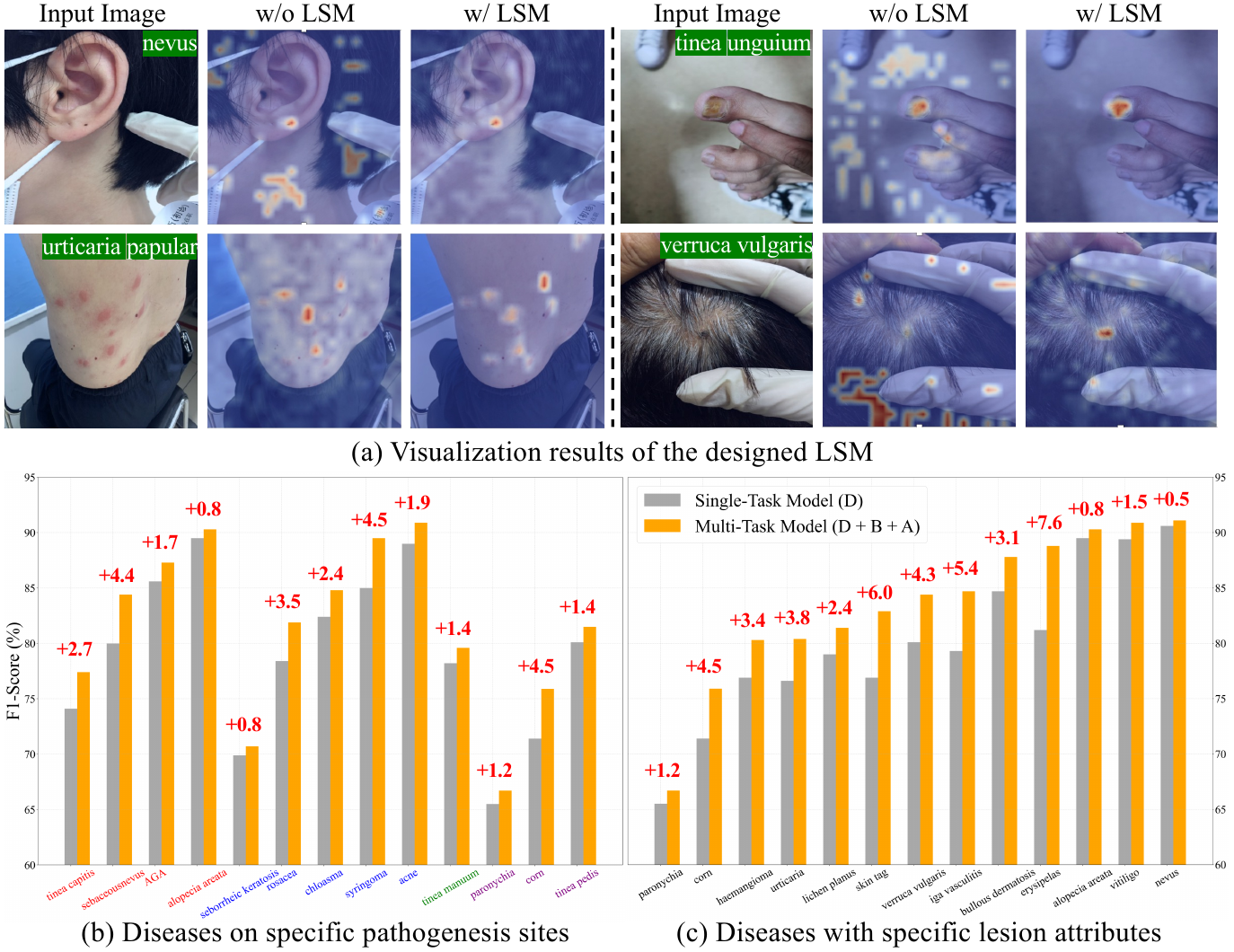}
    \caption{Comparative results on Derm-49 dataset. \textcolor{red}{Red}, \textcolor{blue}{blue}, \textcolor{green}{green}, and \textcolor{purple}{purple} fonts denote diseases on heads, faces, hands, and feet (best viewed in color).}
    \label{fig:figure4}
\end{figure}
\subsubsection{Ablation Study}
 The experiment is conducted based on ViT-B/16 and the results are reported in Table \ref{tab:ablation1}. 
(1) \textbf{LSM:} Quantitative results demonstrate that the designed LSM yields 1.6\% improvement in accuracy. Qualitative results are shown in Fig. \ref{fig:figure4}(a), which depicts the attention maps obtained from the last transformer layer. Without LSM, vision transformers would struggle of localizing lesions and produce noisy attention maps. With LSM, the attention maps are more discriminative and lesions are localized precisely, regardless of variations in terms of scale and distribution. 
(2) \textbf{Multi-task learning:} The models are trained with a shared backbone and different combinations of task-specific heads. The results show that multi-task learning (D+B+A) increases the F1-score from 77.8 to 79.1.
(3) \textbf{CIM:} Quantitative results show that the presented CIM can further improve the F1-score of diseases to 79.5. Notably, the $p$-value of 1.03e-05 ($< 0.01$) is calculated by comparing the results of 5-fold cross-validation with the baseline, illustrating the significance of our model.
In particular, the representation with fused features of body parts and attributes can improve the recognition performance of diseases. As shown in Fig. \ref{fig:figure4}(b) and (c), statistics show that the classification performance of these diseases is improved by the multi-task learning strategy and CIM. For instance, rosacea and tinea versicolor share the same attributes of macule and papular, but rosacea typically affects the face. By fusing the representation of body parts, the F1-score of rosacea is increased by 4.5 \%. Similarly, our model improves the recognition accuracy of diseases with distinctive lesion attributes such as skin tags, urticaria, etc. Meanwhile, the extra information about body parts and attributes improves the interpretability of diagnoses.
\vspace{-2em}

\begin{table}[t]
    \centering
    \setlength{\tabcolsep}{0.35mm}
    \caption{Comparison to state-of-the-art methods on Derm-49 dataset (top) and two public benchmarks: SD-198 dataset (mid), PAD-UFES-20 dataset (bottom).} 
    \begin{tabular}{c|ccccc}
    \hline
    Datasets & Methods    &   F1-score (\%)     & Precision (\%)           & Recall (\%)          & Accuracy (\%)   \\ 
    \hline
    \multirow{4}{*}{\rotatebox{90}{Derm-49}} & DX \cite{jalaboi2023dermx} & 72.6\textpm2.3 & 73.7\textpm0.6 & 72.2\textpm3.1 & 73.4\textpm0.7   \\
     & ViT-Base \cite{dosovitskiy2020image}  &  75.9\textpm0.8 & 80.6\textpm0.7 & 72.9\textpm0.9 & 80.4\textpm0.4     \\
     & Swin-Base \cite{liu2021Swin} &  76.6\textpm0.6 & 83.5\textpm1.1 & 71.0\textpm0.9 & 80.6\textpm0.5   \\
    \cline{2-6}
     & \textbf{DermImitFomer}  &\textbf{78.8\textpm0.5}   &\textbf{83.5\textpm0.6}     &\textbf{74.6\textpm1.1}  &\textbf{82.6\textpm0.5}   \\
    \hline
    \hline
    \multirow{8}{*}{\rotatebox{90}{SD-198}} & SPBL\cite{yang2019self}      &66.2\textpm1.6  &71.4\textpm1.7   &65.7\textpm1.6   &67.8\textpm1.8   \\
     & Aux-D \cite{2020JPhCS1631a2046X}  &68.0\textpm1.0   &67.9\textpm1.0   &69.2\textpm0.9   & -    \\
     & Dual Stream\cite{DSN}         &70.9\textpm1.2   &73.1\textpm1.4   &69.2\textpm1.1   &71.4\textpm1.1    \\
     & TPC \cite{lei2020class}  &63.2\textpm1.6   &65.6\textpm1.7   &64.7\textpm1.6   & -    \\
     & IASN \cite{IAS}       &68.6\textpm0.7   &71.9\textpm0.8   &70.0\textpm0.9   &70.7\textpm0.8     \\
     & PCCT \cite{chen2022pcct}  &65.2\textpm1.6   &68.4\textpm1.4   &66.0\textpm1.5   & -    \\
    \cline{2-6}
     & \textbf{DermImitFomer-ST}  &\textbf{73.6\textpm2.6}   &\textbf{76.1\textpm2.6}     &\textbf{75.1\textpm2.2}  &\textbf{74.5\textpm2.6}   \\ 
    \hline
    \hline
    \multirow{7}{*}{\rotatebox{90}{PAD-UFES-20}} & PAD \cite{pacheco2020impact}      & 71.0\textpm2.9 & 73.4\textpm2.9 & 70.8\textpm2.8 & 70.7\textpm2.8     \\
     & T-Enc \cite{ou2022deep}      & -              & -              & -              & 61.6\textpm5.1     \\    
    \cline{2-6}
     & ResNet-50 \cite{pacheco2020impact} & 67.8\textpm3.7 & 72.0\textpm4.1 & 67.0\textpm4.1 & 67.1\textpm4.1    \\
     & ViT-Base \cite{dosovitskiy2020image}  & 69.9\textpm1.4 & 69.4\textpm1.5 & 70.4\textpm2.2 & 70.6\textpm1.8    \\
     & Swin-Base \cite{liu2021Swin} & 72.1\textpm2.5 & 72.0\textpm2.9 & 72.7\textpm2.6 &  72.7\textpm2.5  \\
    \cline{2-6}
     & \textbf{DermImitFomer-ST}  &73.6\textpm2.8   &72.8\textpm3.2     &74.4\textpm2.4  &74.4\textpm2.4  \\
     & \textbf{DermImitFomer}  &\textbf{74.5\textpm2.5}   &\textbf{73.9\textpm2.9}     &\textbf{75.0\textpm2.1}  &\textbf{75.0\textpm2.1}   \\
    \hline
    \end{tabular}
\label{tab:sd198}
\end{table}

\subsubsection{Results}
To evaluate the effectiveness of our proposed DermImitFormer, we conduct a comparison with various state-of-the-art methods on three different datasets. The results are reported in Table \ref{tab:sd198}.
(1) \textbf{Derm-49:} Compared with other state-of-the-art approaches, our proposed DermImitFormer achieves the leading classification performance in our established dataset with the 5-fold cross-validation splits.
(2) \textbf{SD-198:} Since the dataset does not contain labels of lesion attributes and body parts, the proposed DermImitFormer in Single-Task mode (w/o CIM) is implemented in the experiment. The result is based on the provided 5-fold cross-validation splits. Quantitative results in Table \ref{tab:sd198}(mid) demonstrate that our proposed DermImitFormer-ST achieves state-of-the-art classification performance. In contrast to other approaches, our model can precisely localize more discriminative lesion regions and thus has superior classification accuracy.
(3) \textbf{PAD-UFES-20:} The dataset contains labels of diseases and body parts. Thus, the proposed DermImitFormer with different modes is evaluated in the experiment by the 5-fold cross-validation splits. Quantitative results in Table \ref{tab:sd198} (bottom) demonstrate that our proposed model outperforms the CNN-based \cite{pacheco2020impact,ou2022deep}, and transformer-based methods \cite{dosovitskiy2020image,liu2021Swin}, achieving the state-of-the-art classification performance. In particular, the performance of DermImitFormer is better than that of DermImitFormer-ST in Single-Task mode (w/o CIM), which further indicates the effectiveness of the multi-task learning strategy and CIM.
\vspace{-1em}
\section{Conclusion}
\vspace{-0.5em}
In this work, DermImitFormer, a multi-task model, has been proposed to better utilize dermatologists' domain knowledge by mimicking their subjective diagnostic procedures.  Extensive experiments demonstrate that our approach achieves state-of-the-art recognition performance in two public benchmarks and a large-scale in-house dataset, which highlights the potential of our approach to be employed in real clinical environments and showcases the value of leveraging domain knowledge in the development of machine learning models.

\subsubsection{Acknowledgement}
This work was supported by National Key R\&D  Program of China (2020YFC2008703) and the Project of Intelligent Management Software for Multimodal Medical Big Data for New Generation Information Technology, the Ministry of Industry and Information Technology of the People’s Republic of China (TC210804V). 

\bibliographystyle{splncs04}
\bibliography{mybib}
\end{document}